\documentclass{vgtc}                          

\ifpdf
  \pdfoutput=1\relax                   
  \usepackage{graphicx}                
  \DeclareGraphicsExtensions{.pdf,.png,.jpg,.jpeg} 
\else
  \usepackage{graphicx}                
  \DeclareGraphicsExtensions{.eps}     
\fi%

\graphicspath{{figures/}{pictures/}{images/}{./}} 

\usepackage{microtype}                 
\PassOptionsToPackage{warn}{textcomp}  
\usepackage{textcomp}                  
\usepackage{mathptmx}                  
\usepackage{times}                     
\usepackage{cite}                      
\usepackage{tabu}                      
\usepackage{booktabs}                  

\usepackage{epsfig}
\usepackage{amsmath}
\usepackage{amssymb}
\usepackage{comment}
\usepackage{algorithm}
\usepackage{algpseudocode}
\usepackage{enumitem}
\usepackage{float}
\usepackage{subfig}
\usepackage{tabularx}
\usepackage{multirow}
\usepackage{xspace}
\usepackage{xcolor}

\onlineid{0}

\vgtccategory{Research}

\vgtcinsertpkg

\def\modelname{IPVNet} 
\newcolumntype{Y}{>{\centering\arraybackslash}X}

\makeatletter \DeclareRobustCommand\onedot{\futurelet\@let@token\@onedot}
\def\@onedot{\ifx\@let@token.\else.\null\fi\xspace}
\def\eg{\emph{e.g}\onedot} 
\def\ie{\emph{i.e}\onedot}

\def\etal{\emph{et al}\onedot}
\makeatother

\DeclareMathAlphabet{\mathcal}{OMS}{cmsy}{m}{n}

\title{Automated Reconstruction of 3D Open Surfaces \\ from Sparse Point Clouds}

\author{Mohammad Samiul Arshad\thanks{e-mail: mohammadsamiul.arshad@mavs.uta.edu}\\ %
        \scriptsize The University of Texas at Arlington %
\and William J. Beksi\thanks{e-mail: william.beksi@uta.edu}\\ %
     \scriptsize The University of Texas at Arlington}

\teaser{
  \centering
  \subfloat[]{\includegraphics[width=0.49\linewidth, 
  height=0.18\textheight]{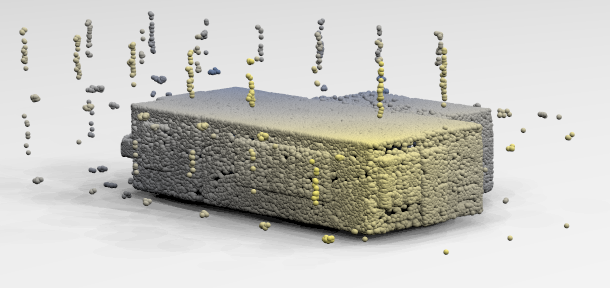}}
  \subfloat[]{\includegraphics[width=0.49\linewidth, 
  height=0.18\textheight]{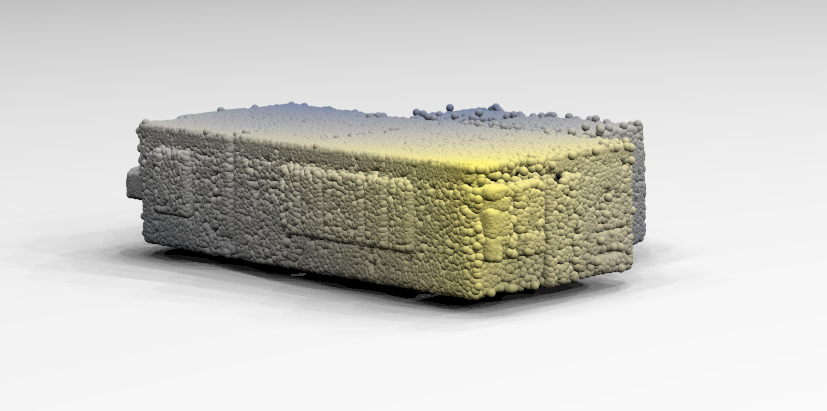}}
  \caption{An outside view of a dense 3D reconstructed scene from the Gibson
  Environment \cite{xia2018gibson} dataset using (a) NDF
  \cite{chibane2020neural}, and (b) our proposed approach. Note that our method
  produces significantly less outliers.}
  \label{fig:short_result}
}

\abstract{
Real-world 3D data may contain intricate details defined by salient surface
gaps. Automated reconstruction of these open surfaces (\eg, non-watertight
meshes) is a challenging problem for environment synthesis in mixed reality
applications. Current learning-based implicit techniques can achieve high
fidelity on closed-surface reconstruction. However, their dependence on the
distinction between the \textit{inside} and \textit{outside} of a surface makes
them incapable of reconstructing open surfaces. Recently, a new class of
implicit functions have shown promise in reconstructing open surfaces by
regressing an unsigned distance field. Yet, these methods rely on a
\textit{discretized representation} of the raw data, which loses important
surface details and can lead to outliers in the reconstruction. We propose
\modelname, a learning-based implicit model that predicts the unsigned distance
between a surface and a query point in 3D space by leveraging both raw point
cloud data and its discretized voxel counterpart. Experiments on synthetic and
real-world public datasets demonstrates that \modelname\ \textit{outperforms}
the state of the art while producing \textit{far fewer} outliers in the
reconstruction.
} 

\CCScatlist{
  \CCScatTwelve{Automated/semi-automated reconstruction}{Open-surface 
  reconstruction}{Unsigned distance field}{}
}

\begin{document}


\firstsection{Introduction}

\maketitle

Capturing detailed point cloud data from the real world is a difficult and
expensive task. Moreover, due to the limitations of 3D sensor technologies (\eg,
LiDAR, RGB-D, etc.), data can be sparse (\ie, missing details) and incomplete
(\ie, noisy with holes and outliers) \cite{guerrero2018pcpnet,li2019pu}.
Automated reconstruction of the missing parts and the reintroduction of surface
details is not a trivial task. Researchers have looked into a myriad of ways
\cite{hoppe1992surface,amenta1998new,amenta2000simple,amenta2001power,
carr2001reconstruction,ohtake2005multi,kazhdan2006poisson,yang2007implicit,
manson2008streaming,galvez2008particle,huang2009consolidation,calakli2011ssd,
kazhdan2013screened,tang2018multi,ren2018biorthogonal} to complete 3D data.
Recently, learning-based implicit functions
\cite{park2019deepsdf,mescheder2019occupancy,chen2019learning,genova2020local,
bhatnagar2020combining,chibane2020neural,chibane2020implicit} have become
popular among 3D reconstruction techniques due to their ability to generate data
in arbitrary resolutions.

One set of implicit learning techniques \cite{atzmon2020sal,atzmon2020sald}
makes use of raw point cloud data to learn a signed distance field. Since
traditional convolutions cannot be applied on permutation invariant point
clouds, such methods often depend on linear feature aggregation through a
multilayer perceptron \cite{qi2017pointnet} or they define a dynamic kernel and
perform a neighborhood search to mimic convolutions
\cite{tatarchenko2017octree}. Other implicit methods
\cite{chibane2020implicit,chibane2020neural} discretize the raw point clouds
into voxel grids. However, voxel grids lose information since multiple points
within the boundary of a grid are merged together. Moreover, the computational
costs and memory requirements increase \textit{cubically} using this approach.
Implicit functions that learn an SDF via extraction of a zero level set must
distinguish between the inside and outside of the surface. As a result, the
reconstruction is produced as a closed surface even if the target shape includes
surface gaps \cite{atzmon2020sald}. However, real-world point cloud data can
consist of salient open surfaces. Closing the surface of such data often leads
to the introduction of outliers and lost details.

To reconstruct accurate geometry and preserve surface details, we propose
\modelname, an implicit model that learns a unsigned distance field (UDF) by
accumulating features from raw point clouds and voxel grids jointly to
reconstruct open surfaces. As shown in \autoref{fig:short_result}, our approach
produces significantly less outliers compared to the state of art
\cite{chibane2020neural}. To the best of our knowledge, our work is the first
approach on combining point-voxel features to learn implicit functions. Our key
contributions can be summarized as follows.
\begin{itemize}
  \item We introduce \modelname, a novel approach for implicitly learning from raw
  point cloud and voxel features to automatically reconstruct complex open 
  surfaces.
  \item We develop an inference module that extracts a zero level set from a UDF
  and drastically lowers the amount of outliers in the reconstruction.
  \item We show that \modelname\ outperforms the state-of-the art on both
  synthetic and real-world public datasets. 
\end{itemize}

\section{Related Work}
\label{sec:related_work}
3D reconstruction is a well researched area with a number of different
approaches and algorithms
\cite{hoppe1992surface,amenta1998new,amenta2000simple,amenta2001power,
carr2001reconstruction,ohtake2005multi,kazhdan2006poisson,yang2007implicit,
manson2008streaming,huang2009consolidation,calakli2011ssd,kazhdan2013screened,
tang2018multi,ren2018biorthogonal,galvez2008particle}. In this section, we
review and compare our work with learning-based implicit approaches. For a more
comprehensive review, we refer to contemporary surveys on 3D reconstruction
\cite{berger2017survey,you2020survey}.

\subsection{Implicit Function Learning}
\label{subsec:implicit_function_learning}
Instead of explicitly predicting a surface, implicit feature learning methods
try to either predict if a particular point in 3D space is inside or outside of
a target surface (occupancy), or determine how far the point is from the target
surface (distance). To reconstruct 3D data in arbitrary resolutions and learn a
continuous 3D mapping, Mescheder \etal \cite{mescheder2019occupancy} presented a
network that predicts voxel occupancy. Peng \etal \cite{peng2020convolutional}
improved the occupancy network by incorporating 2D and 3D convolutions. An
encoder-decoder architecture was used by Chen \etal \cite{chen2019learning} to
learn voxel occupancy. Michalkiewicz \etal \cite{michalkiewicz2019deep}
estimates oriented level set to extract 3D surface. Littwin and Wolf
\cite{littwin2019deep} used encoded feature vectors as the network weights to
predict voxel occupancy. Park \etal \cite{park2019deepsdf} introduced DeepSDF,
an encoder-decoder based architecture that predicts a signed distance to the
surface instead of voxel occupancy. Genova \etal \cite{genova2020local} divided
an object's surface into a set of shape elements and used an encoder-decoder to
learn occupancy. Chibane \etal \cite{chibane2020implicit} used 3D feature
tensors to predict voxel occupancy. Rather than transforming point clouds into a
occupancy grid, Atzmon and Lipman used the raw point clouds to learn and predict
an SDF to the target surface in \cite{atzmon2020sal}, and incorporated
derivatives in regression loss to further improve the reconstruction accuracy in
\cite{atzmon2020sald}.

All of the aforementioned works either predict a voxel occupancy or signed
distance value for a given query point, which is inadequate to reconstruct open
surfaces. To address this problem, we task \modelname\ to learn an unsigned
distance field. Prior to our work, Chibane \etal \cite{chibane2020neural}
predicted a UDF from an input voxel occupancy map. A similar technique to learn
a UDF for single-view garment reconstruction was used by Zhao \etal
\cite{zhao2021learning}. Venkatesh \etal \cite{venkatesh2021deep} proposed a
closest surface point representation to reconstruct both open and close
surfaces. However, preceding work on UDFs often discretize the raw point cloud
data into voxel grids, which results in lost surface details. In contrast, we
make use of the raw point cloud \textit{jointly} with voxel occupancy, thus
enabling us to accumulate improved features and reconstruct finer details with
less outliers.

\subsection{Learning from Points and Voxels}
\label{subsec:point_voxel}
Recently, fusion between features extracted from point cloud and voxel
representations has shown to improve the performance of 3D computer vision
methods. Liu \etal \cite{liu2019point} introduced PVCNN to perform
classification and segmentation by extracting features from both point clouds
and voxel grids via voxelization and de-voxelization.  Fusion between voxel and
point features for 3D classification was used by Li \etal \cite{li2019mvf}. Shi
\etal \cite{shi2020pv} gathered multi-scale voxel features which were combined
into keypoint features from a point cloud for object detection and further
improved the results by incorporating local vector pooling in \cite{shi2021pv}.
Using PVCNN as a backbone, Zhu \etal \cite{zhu20123d} performed shape completion
and generation through point-voxel diffusion. Point-voxel fusion to detect 3D
objects was used by Cui \etal \cite{cui2020pvf} and Tang \etal
\cite{tang2020searching} learned a 3D model via sparse point-voxel convolution.

Noh \etal \cite{noh2021hvpr} accumulated point-voxel features in a single
representation for 3D object detection. PVT, a transformer-based architecture
that learns from point-voxel features for point cloud segmentation was
introduced by Zhang \etal \cite{zhang2021pvt}. Wei \etal \cite{wei2021pv} used
point-voxel correlation for scene flow estimation and Li \etal
\cite{li2021improved} used point-voxel convolution for 3D object detection. Xu
\etal \cite{xu2021rpvnet} introduced RPVNet for point cloud segmentation via
point-voxel fusion. Cherenkova \etal \cite{cherenkova2020pvdeconv} utilized
point-voxel deconvolution for point cloud encoding/decoding. In contrast to the
preceding research, we focus on applying point-voxel fusion to the task of
learning implicit functions and open-surface reconstruction. To the best of our
knowledge, our approach is the first attempt to understand the effectiveness of
point-voxel fusion on these tasks.

\section{Implicit Learning with Point-Voxel Features}
\label{sec:implicit_learning_with_point-voxel_features}
\begin{figure}[t]
\centering
\includegraphics[width=0.99\linewidth,height=0.15\textheight]{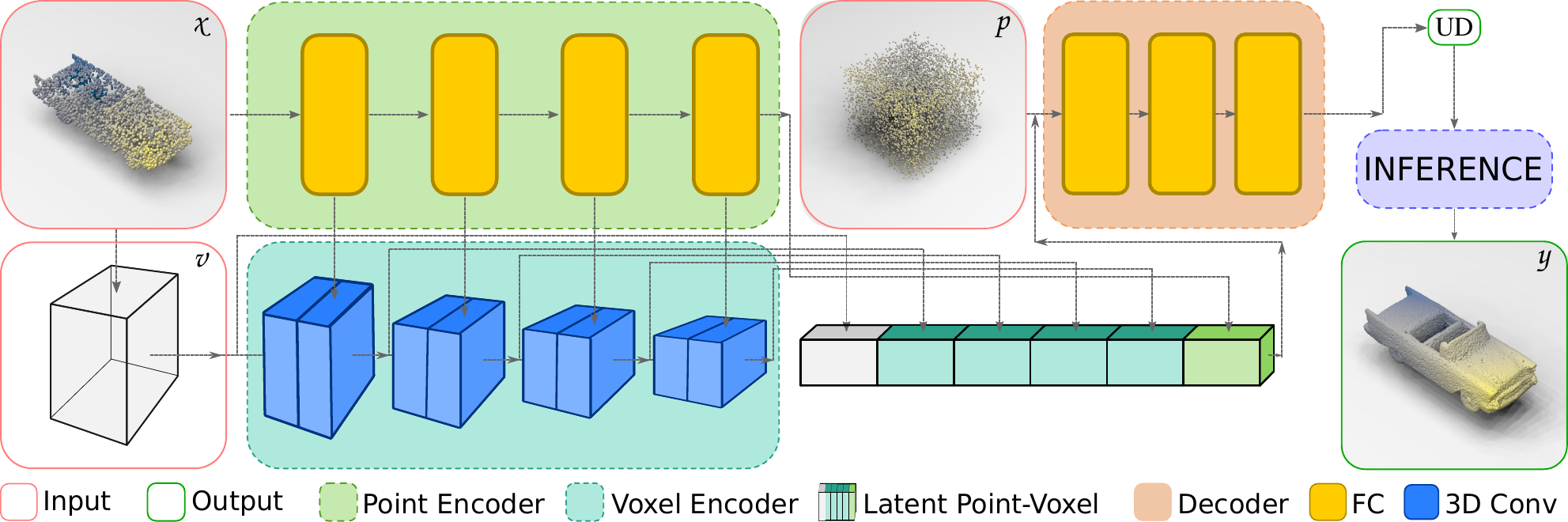}
\caption{Given a sparse point cloud $x \in \mathcal{X}$ of an object, we use a
novel encoding scheme to extract and aggregate point-voxel features from both
the raw point cloud $(x)$ and the voxel occupancy $(v)$. From the accumulated
features, a decoder module regresses the unsigned distance $UD(p,S)$ from
query point $p$ to the surface $S$. By querying the decoder multiple times,
the inference sub-module can reconstruct the surface of any target shape.}
\label{fig:model}
\end{figure}

An overview of our network is presented in Fig.~\ref{fig:model}. Given a sparse
point cloud $x \in \mathcal{X} \subset \mathbb{R}^{N \times 3}$ of an object, we
use a novel encoding scheme to extract and aggregate point-voxel features from
both the raw point cloud $(x)$ and the voxel occupancy $(v)$. From the
accumulated features, a decoder module regresses the unsigned distance $UD(p,S)$
from any query point $p$ to the surface $S$. In the following subsections we
describe the elements of our approach. 

\subsection{Point-Voxel Features}
\label{subsec:point_encoding}
To extract a set of multilevel features from a point cloud $x$, we define a neural
function
\begin{equation}
  \Theta(x) := (z_x^1, \ldots, z_x^j)\: | \: \Theta \colon \mathbb{R}^{N \times 3} 
  \rightarrow \mathbb{Z},
\end{equation}
where $z_x \in \mathbb{Z} \subset \mathbb{R}$ corresponds to the extracted
feature vector from the raw point cloud $x$, and $j$ is the total number of
layers in $\Theta$. ReLu \cite{nair2010rectified} non-linearity is
used for all layers except the output layer of the point encoder.

Instead of limiting the encoded features to a single dimensional vector, a voxel
representation allows for the construction of a multidimensional latent matrix.
However, such an encoding scheme requires the input point cloud $x$ to be
discretized into a voxel grid $v$, \ie, $x \approx v \colon \mathbb{R}^{N \times
3} \rightarrow \mathbb{R}^{M \times M \times M}$ where $M \in \mathbb{N}$ is the
grid resolution. Due to the discretization process, voxel grids lose information
since multiple points may lie within the same voxel. To reintroduce lost
details, we combine voxel features with point features $z_x$.

Let $\Phi \colon \mathbb{R}^{M \times M \times M} \rightarrow \mathbb{Z}^{M \times
M \times M}$ be a neural function that encodes the combined point-voxel features
into a set of multidimensional feature grid $z_{xv}$ of monotonically decreasing
dimension. Then,
\begin{equation}
  \Phi(v \odot \Theta(x)) := (z_{xv}^{k \times k \times k}, \ldots, z_{xv}^{l
  \times l \times l}),
\end{equation}
where $k, l \in \mathbb{N}$ represents the dimensional upper and lower bound of
the feature grid $(M > k \gg l > 1)$, the subscript ${xv}$ denotes the
dependency on both points and voxels. Similar to its point counterpart, the
voxel encoder is more directed towards local details at the early stages.
However, as the dimensionality is reduced and the receptive field grows larger,
the aim shifts to the global structure. ReLu is utilized to ensure non-linearity
and batch normalization \cite{ioffe2015batch} provides stability while training.
The latent point ($z_x^j$) from the point encoder, along with multidimensional
features ($z_{xv}$) from the point-voxel encoder and the discretized voxel grid
($v$), are then used to construct the latent point-voxel
\begin{equation}
  z = \{z_x^j, \Phi(v \odot \Theta(x)), v\}.
\end{equation}

\subsection{Implicit Decoding}
\label{subsec:implicit_decoding}
Given a query point $p \in \mathbb{R}^3$, a set of deep features $F_p$ is
sampled from the latent point-voxel features $z$ via a spatial grid sampling  
\cite{jaderberg2015spatial} function $\Omega$. Specifically,
\begin{equation}
  \Omega(z, p) := (F_p^1 \times \dots \times F_p^{n}),
\end{equation}
where $n = \lvert z \rvert$. Similar to \cite{chibane2020implicit}, we extract
features from a neighborhood of distance $d \in \mathbb{R}$ along the Cartesian
axes centered at $p$ to obtain rich features. More formally,
\begin{equation}
  p := \{p + q \cdot c_i \cdot d\} \in \mathbb{R}^3 \; | \; q \in \{1,0,-1\}, i 
  \in \{1,2,3\},
\end{equation}
where $c_i \in \mathbb{R}^3$ is the $ith$ Cartesian axis unit vector. We define
a neural function $\Psi$ that regresses the unsigned distance to the surface $S$
of $x$ from the deep features ($F_p$). Concretely,
\begin{equation}
  \Psi(F_p^1,\dots,F_p^{n}) \approxeq UD(p,S) \: | \: \Psi \colon \mathbb{Z} 
  \rightarrow \mathbb{R}_+,
\end{equation}
where $UD(\cdot)$ is a function that returns the unsigned distance from $p$ to
the ground-truth surface $S$ for any $p \in \mathbb{R}^3$. Therefore, the
implicit decoder to regress the unsigned distance at a given query point $p$
is defined as
\begin{equation}
  f_x(z,p) := (\Omega \circ \Psi)(p) \: | \: f_x \colon \mathbb{Z} \times 
  \mathbb{R}^3 \rightarrow \mathbb{R}_+.
\end{equation}

\subsection{Training}
\label{subsec:implicit_learning}
\modelname\ requires a pair $\{X_i, S_i\}^T_{i=1}$ associated with input $X_i$
and corresponding ground-truth surface $S_i$ for implicit learning.
Parameterized by the neural parameter $w$, the point-encoder, voxel-encoder,
and decoder are jointly trained with a mini-batch loss
\begin{equation}
  \mathtt{L}_\mathtt{B} := \Sigma_{x \in B} \Sigma_{p \in 
  \mathtt{P}}|\min(f_x^w(p), \delta) - \min(UD(p,S_x), \delta)|,
\end{equation}
where $\mathtt{B}$ is a mini-batch of input and $\mathtt{P} \in 
\mathbb{R}^{Q \times 3}$
is a set of query points within distance $\delta$ of $S_i$. Similar to
\cite{chibane2020neural}, we use a clamped distance $0 < \delta < 10$ (cm) to
improve the models capacity to represent the vicinity of the surface
accurately.

\subsection{Surface Inference}
\label{subsec:surface_inference}
Analogous to \cite{chibane2020neural}, we use an iterative strategy to extract
surface points from $f_x$. More specifically, given a perfect approximator
$f_x(p)$ of the true unsigned distance $UD(p,S_i)$, the projection of $p$ onto
the surface $S_i$ can be obtained by
\begin{equation}
  q := p - f_x(p) \cdot \nabla_pf_x(p), q \in S_i \subset \mathbb{R}^d, \forall
  p \in \mathbb{R}^d/C.
  \label{eq:udf_surface_point}
\end{equation}
In \eqref{eq:udf_surface_point}, $C$ is the cut locus \cite{wolter1993cut}, \ie,
a set of points that are equidistant to at least two surface points. The
negative gradient indicates the direction of the fastest decrease in distance.
In addition, we can move a distance of $f_x(p)$ to reach $q$ if the norm of the
gradient is one. By projecting a point multiple times via
\eqref{eq:udf_surface_point}, the inaccuracies due to $f_x(p)$ being an
imperfect approximator can be reduced. Furthermore, filtering the projected
points to a maximum distance threshold ($max\_thresh$) and re-projecting them
onto the surface after displacement by $d \sim \mathcal{N}(0, \delta /3)$ can
ensure higher point density within a maximum distance ($\delta$).

Instead of uniformly sampling query points within the bounding box of $S_i$
\cite{chibane2020neural}, we use the input points $X_i \in \mathbb{R}^3$ as
guidance for the query points. In particular, we apply a random uniform jitter
$\mathtt{J}_a^b \in \mathbb{R}^3$ within bounds $a$ and $b$ to displace the
input points $X_i$. Due to the inclusion of point features in learning, this
procedure allows our model to infer more accurate surface points while
restricting the number of outliers. The details of the inference
procedure are provided in Alg.~\ref{alg:inference}.

\begin{algorithm}[h]
\caption{Surface Point Inference}
\label{alg:inference}
  \begin{algorithmic}[1]
  \Procedure{Inference}{$\mathtt{X}$}
  \State $np$ $\leftarrow$ Total number of projections
  \State $T$ $\leftarrow$ Maximum threshold distance
  \State $R$ $\leftarrow$ Output resolution of the point cloud
  \State $\mathtt{J} \leftarrow m$ points from $\mathcal{U}(a, b)$
  \State $\mathtt{P}_{init} \leftarrow \{x + j\}, \forall x \in \mathtt{X}, 
  \forall j \in \mathtt{J}$
  \For{$i = 1$ to $np$}
    \State $p \leftarrow p - f_x(p) \cdot \frac{\nabla_p f_x(p)}{||\nabla_p 
    f_x(p)||}, \forall p \in \mathtt{P}_{init}$
  \EndFor
  \State $\mathtt{P}_{filtered} \leftarrow \{p \in 
  \mathtt{P}_{init}\,|\,f_x(p) < T\}$
  \State $\mathtt{P}_{filtered}$: draw $R$ points with 
  replacement
  \State $\mathtt{P}_{filtered} \leftarrow \{p + d\}\,|\,p \in 
  \mathtt{P}_{filtered}, d \sim \mathcal{N}(0, \delta /3)$
  \For{$i = 1$ to $np$}
      \State $p \leftarrow p - f_x(p) \cdot \frac{\nabla_p f_x(p)}{||\nabla_p 
      f_x(p)||}, \forall p \in \mathtt{P}_{filtered}$
  \EndFor
  \State \Return $\{p \in \mathtt{P}_{filtered}\,|\,f_x(p) < T\}$
  \EndProcedure
  \end{algorithmic}
\end{algorithm}


\section{Experiments}
\label{sec:experiments}
To validate the performance of \modelname\ we concentrate on the task of 3D
object and scene reconstruction from sparse point clouds. In this section, we
present the details of the experimental setup and provide an analysis of our
results.

\captionsetup[subfigure]{position=top, labelformat=empty, justification=centering}
\begin{figure}[ht]
\centering
\subfloat[Input]
  {\includegraphics[width=0.19\linewidth]{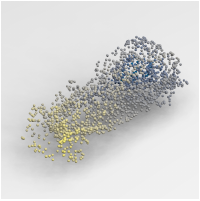}}
   \hspace{0.05pt}
\subfloat[NDF]
  {\includegraphics[width=0.19\linewidth]{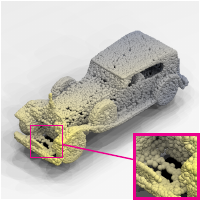}}
   \hspace{0.05pt}
\subfloat[\modelname]
  {\includegraphics[width=0.19\linewidth]{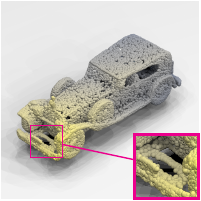}}
   \hspace{0.05pt}
\subfloat[\modelname\ (Inner View)]
  {\includegraphics[width=0.19\linewidth]{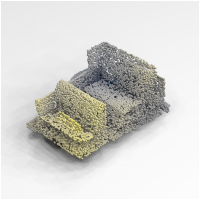}}
   \hspace{0.05pt}
\subfloat[GT]
  {\includegraphics[width=0.19\linewidth]{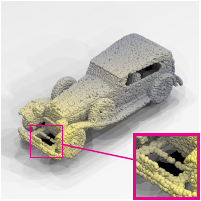}}
\par \vspace{-0.9em}
\subfloat
  {\includegraphics[width=0.19\linewidth]{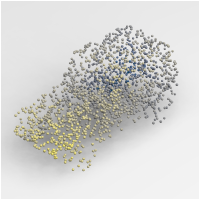}} 
  \hspace{0.05pt}
\subfloat
  {\includegraphics[width=0.19\linewidth]{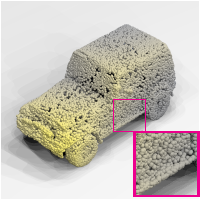}}
   \hspace{0.05pt}
\subfloat
  {\includegraphics[width=0.19\linewidth]{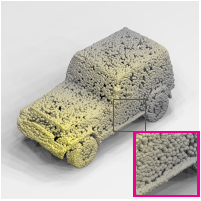}}
   \hspace{0.05pt}
\subfloat
  {\includegraphics[width=0.19\linewidth]{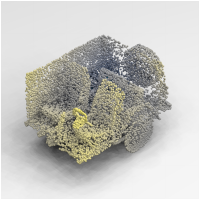}}
   \hspace{0.05pt}
\subfloat{\includegraphics[width=0.19\linewidth]{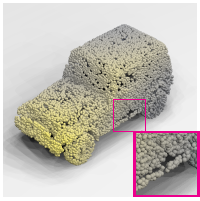}}
\par \vspace{-0.85em}
\subfloat
  {\includegraphics[width=0.19\linewidth]{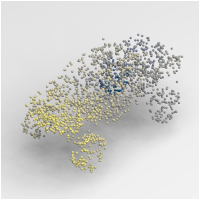}}
   \hspace{0.05pt}
\subfloat
  {\includegraphics[width=0.19\linewidth]{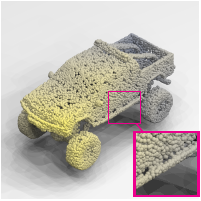}}
   \hspace{0.05pt}
\subfloat
  {\includegraphics[width=0.19\linewidth]{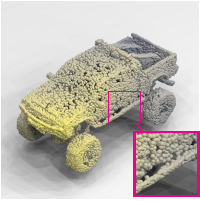}}
   \hspace{0.05pt}
\subfloat
  {\includegraphics[width=0.19\linewidth]{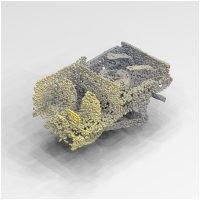}}
   \hspace{0.05pt}
\subfloat
  {\includegraphics[width=0.19\linewidth]{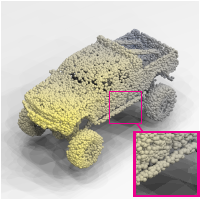}}
\caption{Object reconstruction using NDF \cite{chibane2020neural}, \modelname,
and the ground truth (GT) from the ShapeNet Cars \cite{chang2015shapenet} test
set. \modelname\ performs better on reconstructing thin structures and
preserving small gaps (inset images).}
\label{fig:result_obj_recon}
\end{figure}

\subsection{Baseline and Metric}
We utilize a neural distance field (NDF) \cite{chibane2020neural} as the
baseline to compare the reconstruction quality of \modelname. To the best of
our knowledge, NDF \cite{chibane2020neural} and CSP \cite{venkatesh2021deep}
are the only approaches available for open surface reconstruction. However, due
to the unavailability of the codebase for multi-shape learning via CSP, we
compare our results only against NDF. Note that, to be fair, we do not compare
against methods that are limited to closed surface reconstruction as our main
objective is the reconstruction of open surfaces.  For an unbiased comparison,
we trained an NDF following the directions from \cite{chibane2020neural} on our
train-test split until a minimum validation accuracy was achieved.

To quantitatively measure the reconstruction quality, we use the
\textit{chamfer-}$L_2$ distance (CD) to measure the accuracy and completeness
of the surface. The CD is defined as
\begin{equation}
  d_{CD}(Y, Y_{gt}) = \sum_{i\in Y}^{}\min_{j \in Y_{gt}} {||i-j||}^2 +
  \sum_{j\in Y_{gt}}^{}\min_{i\in Y} {||j-i||}^2,
\end{equation}
where $Y_{gt} \in \mathbb{R}^{\mathcal{O} \times 3}$ is the ground-truth point
cloud, $Y \in \mathbb{R}^{\mathcal{O} \times 3}$ is the reconstructed point
cloud, and $\mathcal{O} \in \mathbb{N}$ is the point density of the ground
truth and the output. In addition, precision and recall are two metrics that
have been extensively used to evaluate 3D reconstruction results. Precision
quantifies the accuracy while recall assesses the completeness of the
reconstruction. For the ground truth $Y_{gt}$ and reconstructed point cloud
$Y$, the precision of an outcome at a threshold $d$ can be calculated as
\begin{equation*}
  P(d) = \sum_{i \in Y}^{}[\min_{j \in Y_{gt}} ||i-j|| < d].
\end{equation*}
Similarly, the recall for a given $d$ may be computed as
\begin{equation*}
R(d) = \sum_{j \in Y_{gt}}^{}[\min_{i \in Y} ||j-i|| < d].
\end{equation*}
The F-score, proposed in \cite{tatarchenko2019single} as a comprehensive
evaluation, combines precision and recall to quantify the overall
reconstruction quality. In detail, the F-score at $d$ is given by
\begin{equation*}
  F(d) = \frac{2 \cdot P(d) \cdot R(d)}{P(d) + R(d)}.
\end{equation*}
An F-score of 1 indicates perfect reconstruction.

\subsection{Object Reconstruction}
Due to the abundance of surface openings, we have chosen the ``Cars" subset
from the ShapeNet \cite{chang2015shapenet} dataset for our object
reconstruction experiment. We used a random split of $70\%$-$10\%$-$20\%$ for
training, validation, and testing, respectively. To prepare the ground truth
and input points we followed the data preparation procedure outlined in
\cite{chibane2020neural}. Additionally, we fixed the output point density
$\mathcal{O}=1$ million to extract a smooth mesh from the point cloud using a
naive algorithm (\eg, \cite{bernardini1999ball}).

To understand the effects of sparse input on the reconstruction quality, we
evaluated \modelname\ and the baseline using an input density of $N \sim
\{300,3000,10000\}$ points while fixing the voxel resolution to $M = 256$. In
contrast to the baseline, \modelname\ can reconstruct thin structures more
accurately and preserve small gaps (see the inset images in
Fig.~\ref{fig:result_obj_recon}) while quantitatively outperforming the
reconstruction with different input densities
(Table~\ref{tab:result_obj_recon}).

\begin{table}[h]
\begin{tabularx}{\columnwidth}{c | Y | Y | Y | Y | Y } \hline
              & \multicolumn{3}{c|}{\textit{Chamfer-}$L_2 \downarrow$} &  
              \multicolumn{2}{c}{$F$-\textit{score} $\uparrow$}\\ 
              \cline{2-6}
              & $ N = 300$          & $N = 3000$        & $N = 10000$ 
              & $d = 0.1\%$         & $d = 0.05\%$ \\ \hline
  NDF         & $1.550$             & $0.324$           & $0.092$ 
              & $0.711$             & $0.460$ \\
  \modelname\ & $\mathbf{1.217}$    & $\mathbf{0.119}$  & $\mathbf{0.068}$ 
              & $\mathbf{0.785}$    & $\mathbf{0.542}$ \\ \hline
\end{tabularx}
\caption{A quantitative comparison between \modelname\ and the baseline (NDF
\cite{chibane2020neural}) on the ShapeNet Cars \cite{chang2015shapenet} dataset
for object reconstruction from different input densities. \modelname\
outperforms the baseline on all input densities. The chamfer-$L_2$ results are
of order $\times 10^{-4}$ and the reconstruction results using an input density
of $N = 10000$ were used to calculate the $F$-\textit{score}.}
\label{tab:result_obj_recon}
\end{table}

\subsection{Real-World Scene Reconstruction}
\begin{figure}[ht]
\centering
\subfloat[Input]
  {\includegraphics[width=0.24\linewidth]{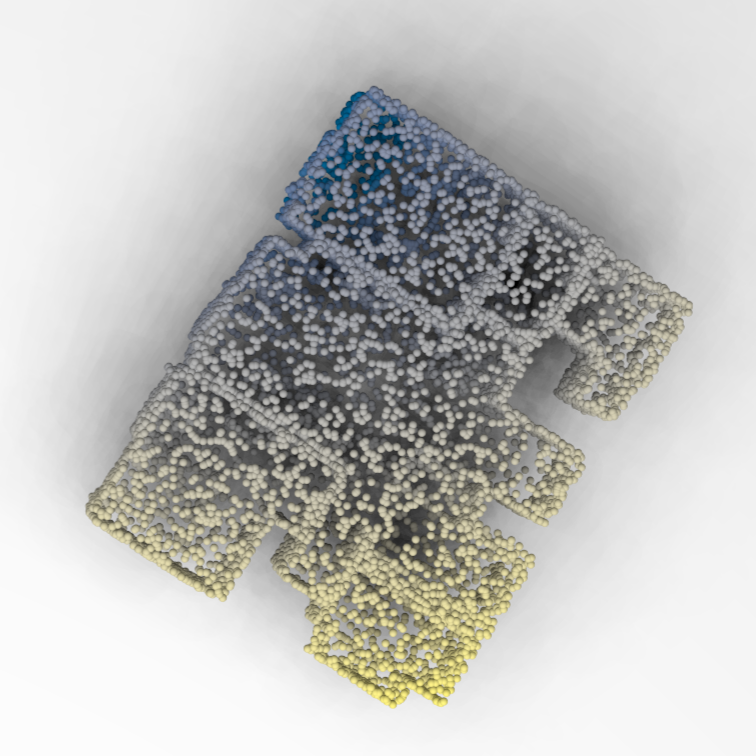}} \hspace{0.05pt}
\subfloat[NDF]
  {\includegraphics[width=0.24\linewidth]{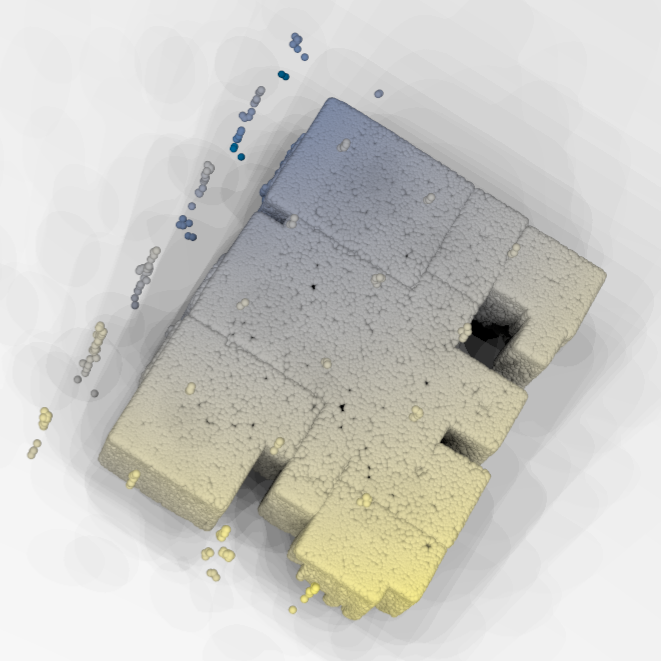}} \hspace{0.05pt}
\subfloat[\modelname]
  {\includegraphics[width=0.24\linewidth]{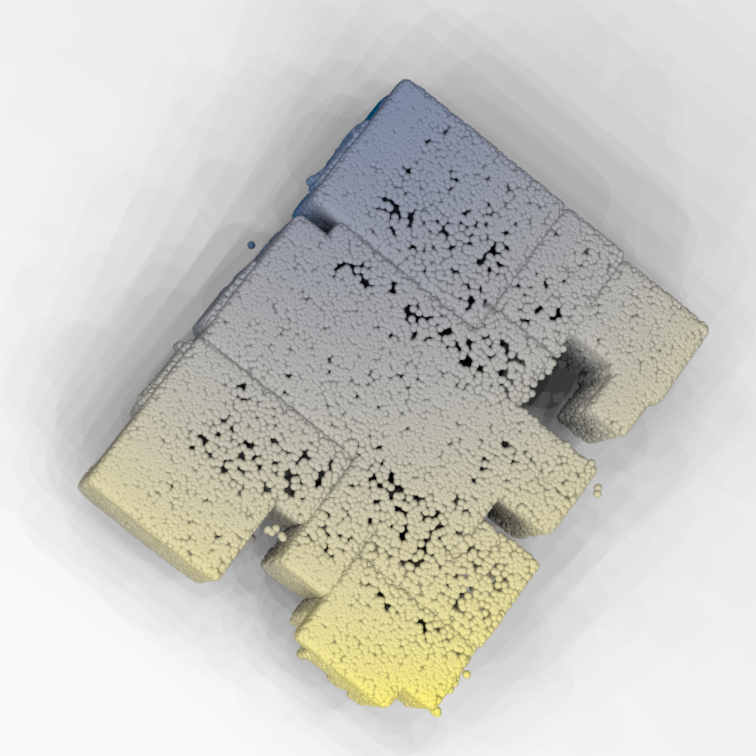}} \hspace{0.05pt}
\subfloat[GT]
  {\includegraphics[width=0.24\linewidth]{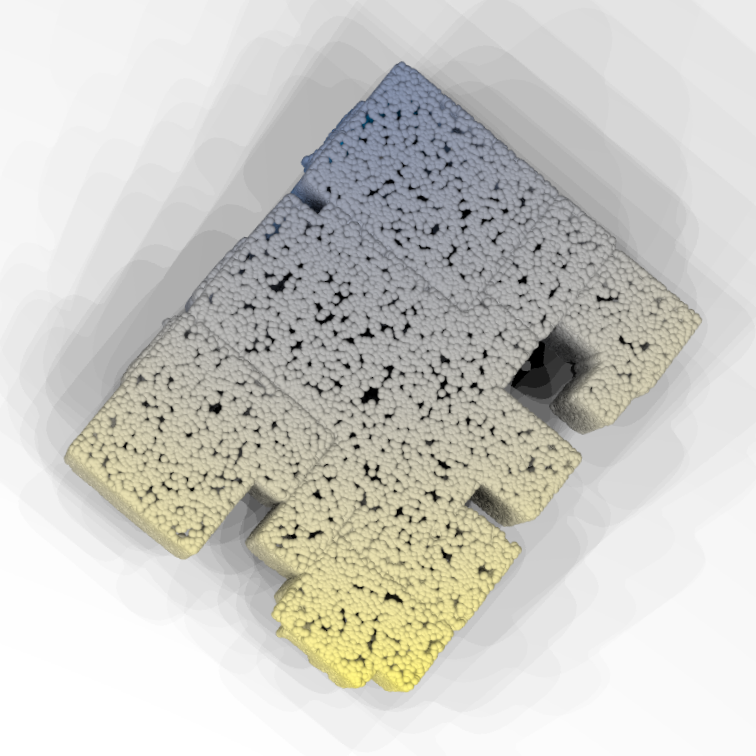}}
\par \vspace{-0.9em}
\subfloat
  {\includegraphics[width=0.24\linewidth]{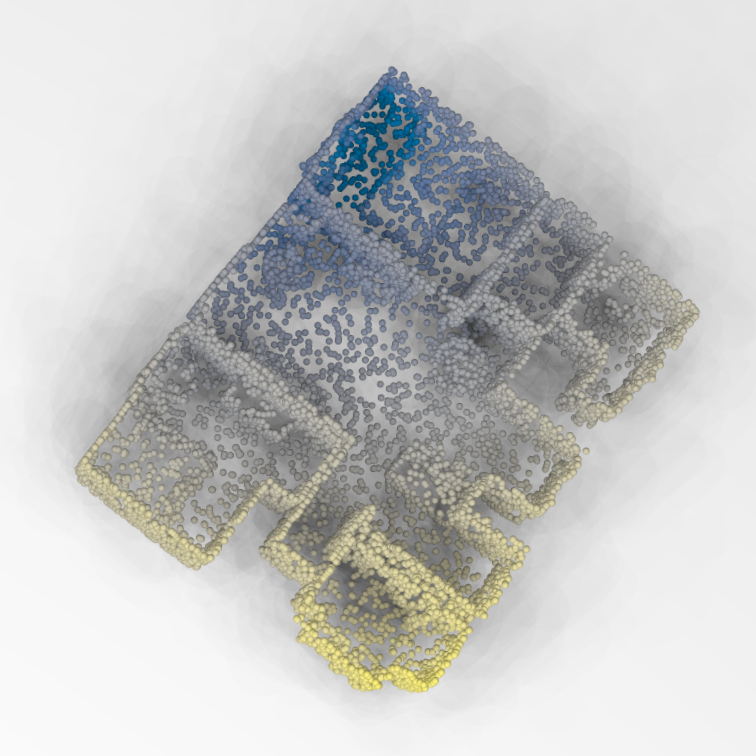}} \hspace{0.05pt}
\subfloat
  {\includegraphics[width=0.24\linewidth]{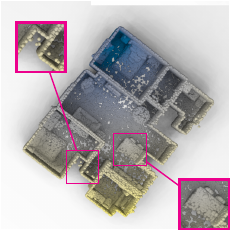}} \hspace{0.05pt}
\subfloat
  {\includegraphics[width=0.24\linewidth]{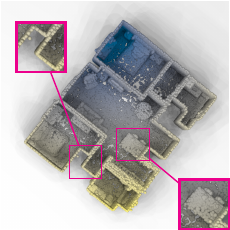}} \hspace{0.05pt}
\subfloat
  {\includegraphics[width=0.24\linewidth]{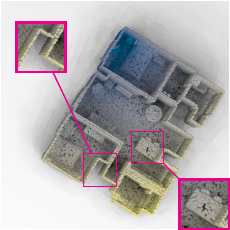}} \hspace{0.05pt}
\par \vspace{-0.9em}
\subfloat
  {\includegraphics[width=0.24\linewidth]{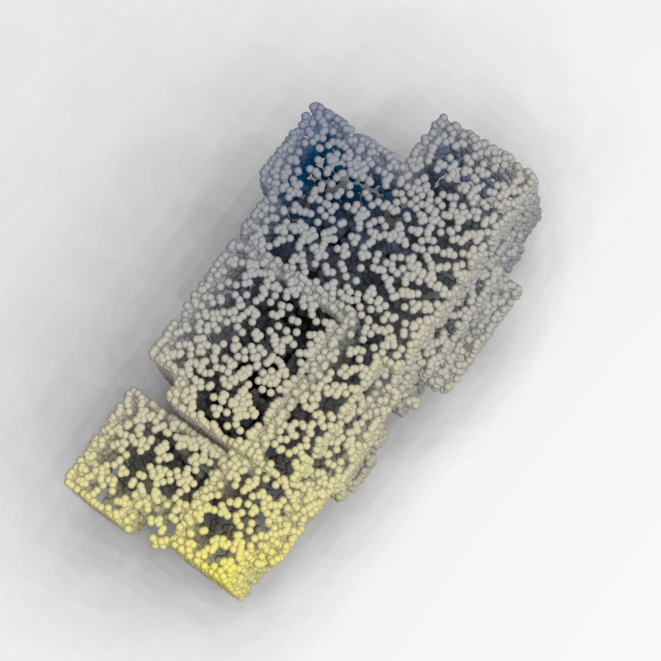}} \hspace{0.05pt}
\subfloat
  {\includegraphics[width=0.24\linewidth]{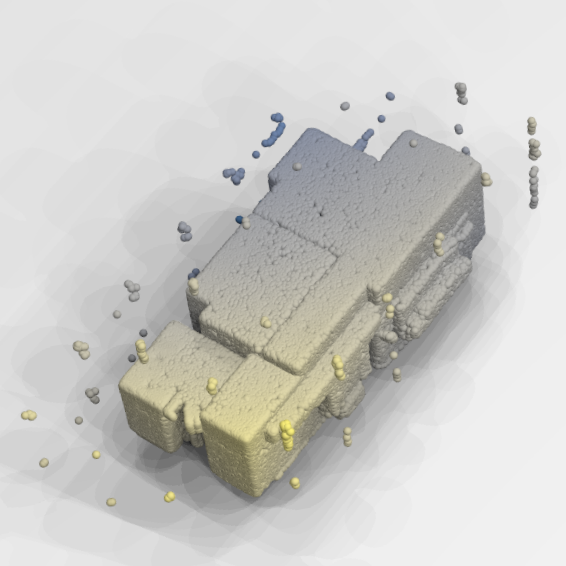}} \hspace{0.05pt}
\subfloat
  {\includegraphics[width=0.24\linewidth]{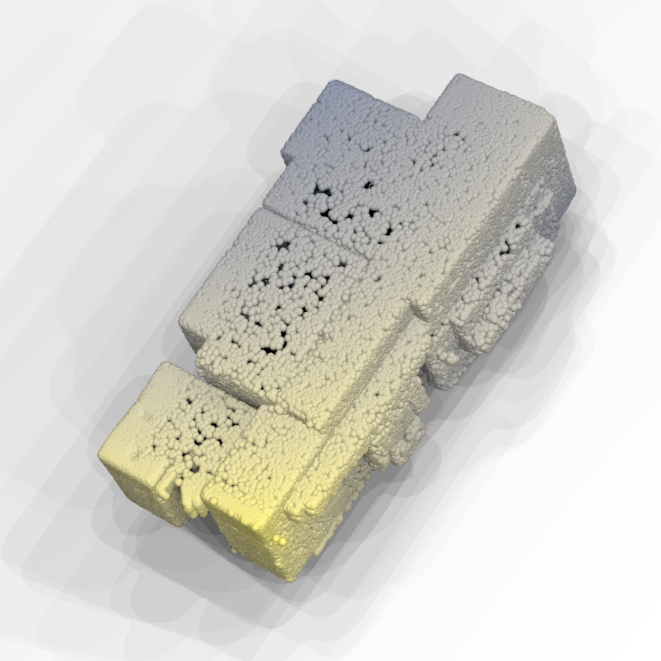}} \hspace{0.05pt}
\subfloat
  {\includegraphics[width=0.24\linewidth]{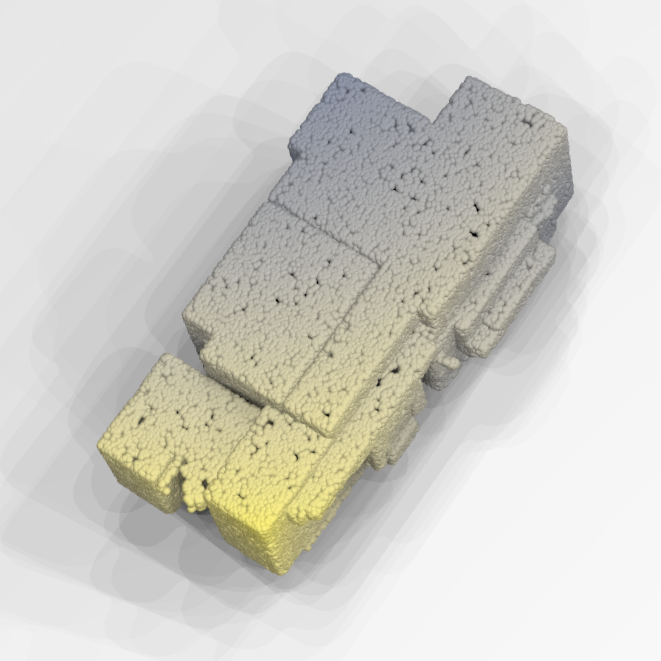}}
\par \vspace{-0.9em}
\subfloat
  {\includegraphics[width=0.24\linewidth]{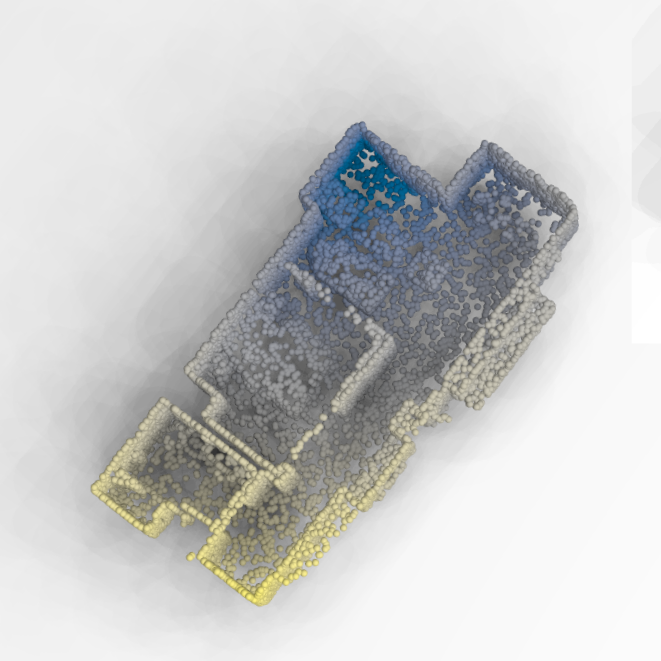}} \hspace{0.05pt}
\subfloat
  {\includegraphics[width=0.24\linewidth]{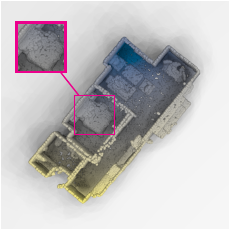}} \hspace{0.05pt}
\subfloat
  {\includegraphics[width=0.24\linewidth]{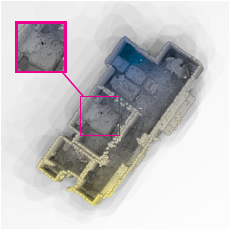}} \hspace{0.05pt}
\subfloat
  {\includegraphics[width=0.24\linewidth]{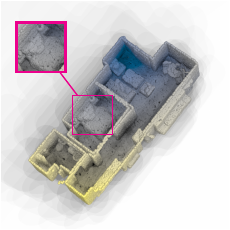}} \hspace{0.05pt}
\par 
\caption{Scene reconstruction on the test set of the Gibson Environment
\cite{xia2018gibson} dataset using NDF \cite{chibane2020neural}, \modelname, and
the respective ground truth (GT). Each odd row represents an outside view of a
scene while the even rows depict inside views. In contrast to the baseline,
\modelname\ produces significantly less outliers (outside view) and improves the
preservation of geometric features (inset images).} 
\label{fig:result_scene_recon}
\end{figure}

We evaluate the reconstruction of complex real-world scenes through the use of
the Gibson Environment dataset \cite{xia2018gibson}. The dataset consists of
RDG-D scans of indoor spaces. A subset of 35 and 100 scenes were prepared
following the procedure from \cite{chibane2020neural} for training and testing
respectively. We used a sliding window scheme and reconstructed the surface
bounded by each window. Since the sliding window may frequently consist of a
very small area of the scene with only few points, we used an output density
five times as large the input density (\ie, $\mathcal{O} = 5 \times N$) to save
time. The grid resolutions were kept fixed at $M = 256$ for both \modelname\ and
the baseline. The reconstruction results are highlighted in
Fig.~\ref{fig:result_scene_recon}. In addition to improving the preservation of
structural details, \modelname\ produces significantly fewer outliers than the
baseline due to the use of point features during training and inference.

\captionsetup[subfigure]{position=top, labelformat=empty, justification=centering}

\section{Conclusion}
\label{sec:conclusion}
In this work we have introduced \modelname, a novel architecture for
automatically reconstructing complex open surfaces. To do this, we make use of
raw point cloud data jointly with voxels to learn local and global features.
Not only have we showed that \modelname\ outperforms the state of the art on
both synthetic and real-world data, but we also demonstrated the effectiveness
of point features on 3D reconstruction through ablation studies. Furthermore,
we developed an inference module that extracts a zero level set from a UDF and
drastically reduces the amount of outliers in the reconstruction. \modelname\
is an important step towards reconstructing open surfaces without losing
details and introducing outliers. We believe that our work will inspire more
research on this topic.

\bibliographystyle{abbrv-doi}

\bibliography{automated_reconstruction_of_3dopen_surfaces_from_sparse_point_clouds}
\end{document}